\documentclass{article}

\usepackage{arxiv}

\usepackage{amssymb}
\usepackage{amsmath}

\usepackage{hyperref}       
\usepackage{url}            
\usepackage{booktabs}       
\usepackage{multirow}
\usepackage{amsfonts}       
\usepackage{nicefrac}       
\usepackage{microtype}      
\usepackage{lipsum}	    	
\usepackage{graphicx}
\usepackage[numbers]{natbib}

\graphicspath{ {pics/} }

\title{A Generative Machine Learning Model for Designing Metal Hydrides Applied to Hydrogen Storage}

\date{}

\usepackage{authblk}

\author[1]{Xiyuan Liu \thanks{Corresponding author: \texttt{liuxyuan@latech.edu}\\
\href{https://doi.org/10.1016/j.ijhydene.2026.153744}{https://doi.org/10.1016/j.ijhydene.2026.153744}}}
\author[1]{Christian Hacker}
\author[2]{Shengnian Wang}
\author[3]{Yuhua Duan}
\affil[1]{Mathematics and Statistics, Louisiana Tech University, Ruston, LA 71272, United States}
\affil[2]{Institute for Micromanufacturing, Louisiana Tech University, Ruston, LA 71272, United States}
\affil[3]{National Energy Technology Laboratory, United States Department of Energy, Pittsburgh, PA 15236, United States}

\begin{document}
\maketitle

\begin{abstract}
Developing new metal hydrides is a critical step toward efficient hydrogen storage in carbon-neutral energy systems. However, existing materials databases, such as the Materials Project, contain a limited number of well-characterized hydrides, which constrains the discovery of optimal candidates. This work presents a framework that integrates causal discovery with a lightweight generative machine learning model to generate novel metal hydride candidates that may not exist in current databases. Using a dataset of 450 samples (270 training, 90 validation, and 90 testing), the model generates 1,000 candidates. After ranking and filtering, six previously unreported chemical formulas and crystal structures are identified, four of which are validated by density functional theory simulations and show strong potential for future experimental investigation. Overall, the proposed framework provides a scalable and time-efficient approach for expanding hydrogen storage datasets and accelerating materials discovery.
\end{abstract}


\section{Introduction}
\label{Introduction}
Hydrogen holds great promise as a transportation and storage solution in the carbon-neutral energy mission. However, the widespread use of hydrogen energy relies on the availability of safe, efficient, and cost-effective storage processes \cite{osman2024advances}. Current hydrogen storage strategies adopt either physical-based (e.g., compression, liquefaction, and cryo-compressed storage) or material-based approaches (e.g., metal hydrides and porous adsorbents), each with distinct strengths and weaknesses. Physical hydrogen storage methods are mature and well-established in industry, but remain quite expensive and dangerous due to their extreme pressure or cryogenic-temperature requirements  \cite{elberry2021large, valenti2016hydrogen, brunner2016cryo}. Material-based storage systems are much safer due to their low operating pressure and temperature, and are more attractive because of their high volumetric hydrogen densities, and reversible storage capabilities \cite{sathe2023furtherance, yang2008material, cousins2019highly, singh2023material}. But problems such as slow kinetics and/or low gravimetric storage capacity remain significant obstacles to their commercialization. Among various material-based hydrogen storage systems, metal hydrides show promising prospects for long-term practice and high-volume applications \cite{drawer2024metal, laurencelle2009experimental}. However, current known metal and alloy hydrides for hydrogen storage are limited, and most of them face similar drawbacks of material-based storage systems, such as low hydrogen storage capacity, slow hydrogen absorption or desorption kinetics, limited cycle life, and other unfavorable thermodynamic properties \cite{bishnoi2024architectural, klopvcivc2023review, xu2024rare}. Despite decades of study, desirable metal hydrides that can safely and efficiently store and release hydrogen are yet to be identified, largely because of the high cost and time-consuming discovery process. When this applies to multinary complex or alloy hydrides that researchers currently focus on, the difficulty increases dramatically for both existing experimental and computational screening tools. It is almost impractical to explore that large number of material candidates and their various combinations using traditional exploration methods, which unfortunately show promising potential to identify appropriate metal hydrides that satisfy all required criteria for hydrogen storage.

With the advancement of machine learning (ML), several alternative exploration techniques have been proposed to accelerate the discovery of new materials. These ML methods can generally be divided into two types: classical ML and Deep Learning (DL) approaches \cite{osman2024advances, franic2025review}.

\subsection{Classical machine learning approach}
Classical ML typically requires less data and is easier to train, making it suitable for experimental design-type datasets where researchers have prior knowledge of the relationships between variables. Examples include formation energy prediction using Gaussian Process Regression \cite{gheytanzadeh2022estimating}, Decision Trees \cite{liu2022key}, and Support Vector Machines (SVM) \cite{wan2021data}. Furthermore, in the field of material generation, Grand Canonical Monte Carlo (GCMC) and Reverse Monte Carlo (RMC) methods are also widely recognized \cite{osman2024advances}.

Although these models provide significant improvements in prediction and interpretation, they typically perform poorly when researchers have limited prior knowledge or when the experimental subject is in an exploratory phase. To address this issue, neural networks (NNs), which are capable of approximating complex nonlinear functions, have been introduced and further developed into deep learning (DL) models.

\subsection{Deep learning model approach}
Classified as an advanced ML approach, models in the class of DL typically achieve higher prediction accuracy at the cost of interpretability and increased model complexity. Recent developments in NNs and deep learning have enabled the use of models such as Diffusion Probabilistic Models, Graph Neural Networks (GNNs), and Recurrent Neural Networks (RNNs), which demonstrate enhanced accuracy and greater potential in predicting material properties, including formation energy \cite{miraki2025probabilistic, choudhary2022recent}.

However, despite their strong predictive performance, these DL models, like classical approaches, operate as conditional estimators (that is, discriminative models) that evaluate existing chemical formulas based on their features. As a result, they cannot generate new material candidates that are absent from the training dataset and therefore cannot directly support the discovery of novel metal hydrides for hydrogen storage applications.

\subsection{Generative machine learning approach}
To address this limitation, as an alternative to discriminative models, researchers have developed generative machine learning approaches such as Variational Autoencoders (VAEs)~\cite{pakornchote2024diffusion, xiecrystal} and Generative Adversarial Networks (GANs)~\cite{long2021constrained} to generate new crystal structures under specified conditions. However, because these generative models estimate the full joint distribution of the variables rather than only conditional distributions, they require significantly larger parameter spaces compared with discriminative models. As a result, these models typically require large training datasets and long training times, making them less practical for exploring chemical space in hydrogen storage and materials science more broadly. This limitation is further amplified by the practical difficulty and cost of collecting large-scale experimental data from laboratory measurements~\cite{xu2023small}.

Therefore, a new framework is urgently needed. First, it should allow practical interpretation of the relationships between features of metal hydrides and identify the most important features to use for generating new chemical compounds. Second, the generative model should be computationally efficient and capable of training on relatively small datasets to generate reasonable new metal hydrides.

\subsection{Causal discovery}
A statistically principled way to address this challenge is to reduce the number of unknown model parameters by limiting the set of variables (i.e., predictors) associated with the target variable. In classical machine learning, this process is referred to as feature selection. However, many feature selection approaches, such as LASSO~\cite{tibshirani1996regression} and Best Subset Selection~\cite{james2013introduction}, rely on linear relationships among variables and assume that predictors are mutually independent or only weakly correlated. These assumptions are often unrealistic in materials science, where variables typically exhibit nonlinear interactions and strong dependencies.

To address these limitations, causal discovery methods are introduced~\cite{spirtes2000causation}. These methods integrate statistical hypothesis testing with graph theory to infer relationships among variables and can be used to identify the Markov blanket of the target variable, representing the subset of features most relevant for prediction.

In this paper, we propose a framework that combines causal discovery with generative ML modeling to produce new hydrogen storage candidates along with their potential crystal structures. After generation, a rule-based method filters out economically or practically infeasible compounds, and the remaining candidates are ranked to identify the most promising metal hydrides. Remarkably, the proposed method trains the generative ML model using only 450 observations  (270 for training, 90 for validation, and 90 for testing samples).

Using this framework, we generate 1,000 candidate materials. After filtering and ranking, six previously unreported alloy hydrides are selected and further evaluated using DFT simulations. Four of these exhibit practical feasibility and favorable hydrogen storage characteristics, indicating strong potential for future experimental validation. In addition, unlike DFT calculations, which typically require access to high-performance computing resources, the proposed machine learning pipeline can be executed on a personal computer equipped with a consumer-grade GPU (NVIDIA RTX 3090), resulting in considerably lower computational cost.

This study introduces several key innovations. First, it presents a causal discovery–guided feature identification strategy that leverages conditional independencies rather than conventional correlation-based feature selection. This approach mitigates the curse of dimensionality, constrains the effective parameter space, and consequently reduces the required training data size. Building on this foundation, the study demonstrates a lightweight generative machine learning framework that achieves strong performance even when trained on an extremely small dataset. Most importantly, the proposed generative model enables the design of novel metal hydrides with explicit crystal structure information. These generated compounds may not exist in current materials databases but have the potential to exhibit enhanced hydrogen storage performance, and the predicted crystal structures can further guide experimental synthesis and computational validation. While the present work focuses on thermodynamic-property-based prediction, extending the framework to incorporate kinetic and dynamic behaviors through time-series analysis, such as Long Short-Term Memory (LSTM)–based ML, represents a promising direction for future research.


The remainder of this paper is organized as follows. Section \ref{RelativeWorks} provides a review and introduction to the methods used in this framework. Section \ref{ProposedMethod} details how these methods are integrated into the proposed framework. Section \ref{Experiments} presents the application of the method to real-world data and analyzes the results. Finally, Section \ref{Conclusion} summarizes the advantages and limitations of the proposed method and discusses potential future research directions.

\section{Related works}\label{RelativeWorks}

As mentioned in the previous section, one major challenge in applying ML to hydrogen storage material discovery is that deep learning and generative models typically require large datasets, which is impractical in this domain. Another challenge is the limited availability of experimentally validated data and the difficulty of expanding datasets without trial-and-error experimentation. Furthermore, due to the complexity of the underlying physics, the available data are often high-dimensional, which introduces the curse of dimensionality and reduces the accuracy and reliability of ML models~\cite{zhou2025machine,xu2023small}.

The curse of dimensionality in ML occurs when too many input variables, especially those unrelated to the response, increase computational complexity without improving predictive performance \cite{james2013introduction}. This issue arises in hydrogen storage research because researchers often have limited understanding of the relationships between chemical features. To avoid the curse of dimensionality and develop a lightweight generative ML model that requires only a small dataset, it is necessary to first identify a few chemical features that are most relevant. This is achieved by defining a hydrogen storage score as an index to evaluate existing compounds and then applying a highly interpretable ML model to identify the most important features related to hydrogen storage.


\subsection{Hydrogen storage score}
A foundational requirement in hydrogen storage research is the ability to quantitatively assess a material's potential. Nations et al. \cite{nations2023metal} introduced a composite metric that combines thermodynamic properties and hydrogen storage capacity into a single score, termed the hydrogen storage score. This metric, referred to as the H Storage Score, is defined as
\begin{equation}\label{eq:HStorageScore}
\text{H Storage Score} = E_{factor}W_{H_2},
\end{equation}
where $W_{H_2}$ is the hydrogen weight fraction obtained from the database and the chemical formula, and $E_{factor}$ is a formation energy-based weighting factor given by
\begin{equation}\label{eq:E_factor}
E_{factor} = \begin{cases}
\sqrt{1 - \left(\frac{|E_{form} + 0.5|}{0.5}\right)^2}, & E_{form} \in [-1, 0] \text{ eV},\\
0, & \text{otherwise},
\end{cases} \end{equation}
with $E_{form}$ representing the formation energy. Equation \eqref{eq:E_factor} focuses on materials with slightly negative formation energies because they offer a good balance that is important for practical hydrogen storage: materials need to be stable enough to hold hydrogen safely, but not too stable so that releasing hydrogen becomes difficult. The Hydrogen Storage Score helps us quickly find materials that are likely to work well, making it a useful tool for selecting the best candidates.

\subsection{Causal discovery}
After establishing a quantitative metric, the next challenge is to identify the chemical and structural features that directly influence the hydrogen storage score. Traditional feature selection methods often identify correlations that do not indicate causation, resulting in misleading conclusions. To address this, our study uses tools from causal inference, a field focused on uncovering cause-and-effect relationships from observational data.

A key framework for causal inference is the Directed Graphical Causal Model (DGCM), introduced by Spirtes in 2000 \cite{neuberg2003causality}. With advances in data collection and the growth of big data analytics, DGCMs have become widely used in areas such as internet service quality assessment, supply chain management, image processing, and Alzheimer’s pathophysiology \cite{meng2020localizing, bo2024root, ye2023bayesian, shen2020challenges}.

Causal discovery is the algorithmic backbone of causal inference, providing techniques to learn DGCM structures directly from data. These algorithms typically fall into two categories: constraint-based and score-based methods \cite{glymour2019review}.  Constraint-based approaches rely on statistical tests of conditional independence to infer graph structure, while score-based methods search for the optimal causal graph by evaluating a scoring criterion over candidate models. In this study, we adopt the Fast Causal Inference (FCI) algorithm \cite{spirtes2013causal}, a constraint-based method that is particularly well-suited for observational data, as it accounts for hidden confounders and indirect causal pathways.

\subsection{Generative model approach to DFT}
While causal discovery helps narrow down the most influential features, generating new candidate materials remains computationally intensive. DFT simulations are the gold standard for evaluating thermodynamic and structural properties \cite{jones2015density}, but they are prohibitively expensive for large-scale screening. Moreover, DFT requires prior knowledge of the chemical formula and crystallographic data, which restricts its use in exploratory material generation.

To address this challenge, researchers have increasingly incorporated machine learning and generative models into the materials discovery process \cite{zhang2018deepcg, kim2024data, bochkarev2024graph}. These models offer rapid approximations that guide more focused DFT analysis, enhancing overall efficiency.

One promising approach in the field of generative modeling is the VAE, which uses deep learning techniques to generate new data samples based on patterns learned from a training dataset. A VAE consists of two neural networks, known as the encoder and the decoder, which work together to transform high-dimensional input data into a structured and lower-dimensional latent space. From this latent space, new and plausible data samples can be generated. Initially developed for image generation tasks, where it has successfully produced realistic variations of existing images \cite{pu2016variational}, the VAE framework has since been adopted for applications in materials science \cite{xiecrystal}. In this context, VAEs can be trained on known materials data to generate novel chemical structures or compositions. This approach offers a computationally efficient, scalable alternative to traditional methods, such as DFT, for exploring candidate materials.

\section{Proposed method}\label{ProposedMethod}
\begin{figure}[ht]
\centering
\includegraphics[width=0.55\textwidth]{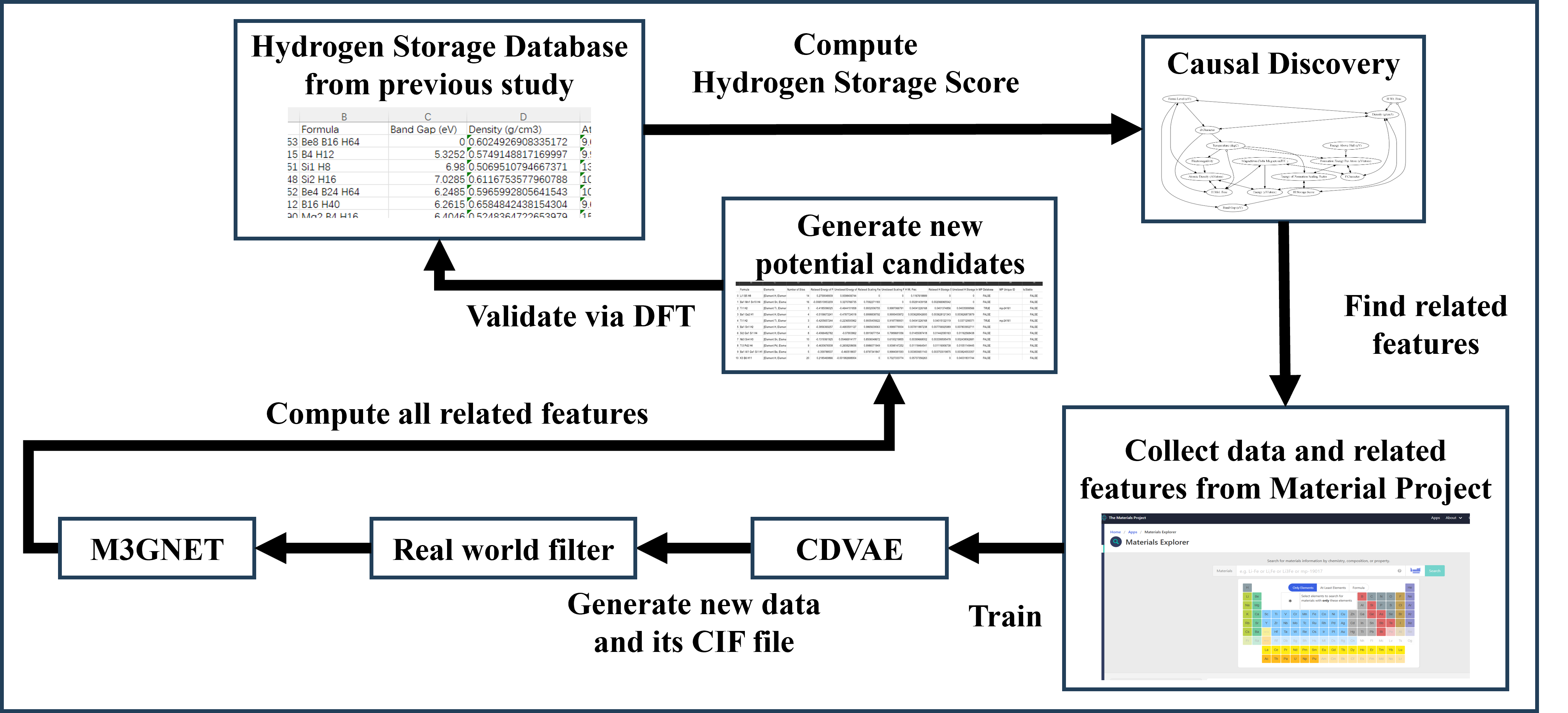}
\caption{Workflow of the proposed method for discovering novel hydrogen storage materials.}
\label{fig:ProposedWork}
\end{figure}

After identifying appropriate evaluation methods using the hydrogen storage score, discovering key features through FCI algorithm, and generating candidates with a VAE-based model, this section describes how these components are integrated into a complete framework (see Fig.~\ref{fig:ProposedWork}).

In this framework, we begin by computing the modified H Storage Score and incorporating it into the Hydrogen Storage Database~\cite{nations2023metal}. This enriched database is then used as input for the FCI algorithm to identify features that are strongly associated with hydrogen storage performance. Next, we extract training data from the Materials Project database~\cite{Jain2013}, including only the features selected by the FCI algorithm along with the computed H Storage Score. These training data are then used to train the CDVAE model.

After training, the CDVAE is used to generate new chemical formulas (i.e., CIF files) along with their predicted H Storage Scores and corresponding crystal structures. These generated structures are then filtered based on practical constraints and a minimum H Storage Score threshold. The filtered candidates are subsequently processed using a pretrained M3GNet model, which relaxes the crystal structures to improve their stability and practical feasibility, and then computes the associated material features.

These candidates, together with their relaxed crystal structures and associated features, are then validated using DFT simulations to ensure stability. Once validated, the new materials and their properties are added to the Hydrogen Storage Database, allowing the causal graph to be updated for the next discovery iteration.


\subsection{Modified hydrogen storage score}
\begin{figure}[ht]
\centering
\includegraphics[width=0.6\textwidth]{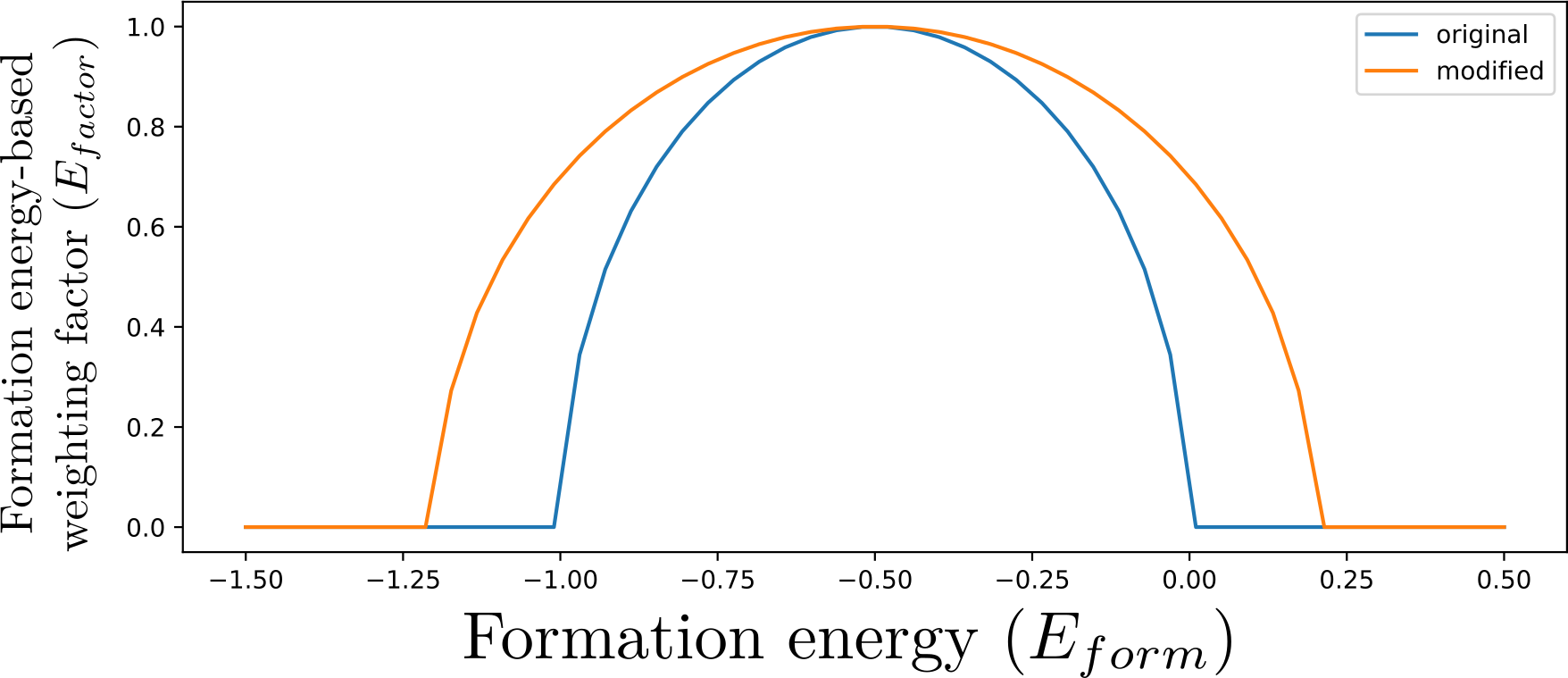}
\caption{Formation energy-based weight factor comparison. The original function is shown in blue, and the modified version is shown in orange.}
\label{fig:E_factor}
\end{figure}

In this study, we modified the calculation formula of the formation energy weighting factor, $E_{factor}$, to enlarge the initial searching database of candidate metal hydrides. Specifically, we extend the non-zero interval for  $E_{form}$ from the original $[-1, 0]$ eV to a wider range of $[-1.2, 0.2]$ eV.
In the Materials Project database, without temperature and/or pressure factors considered, the upper bound of 0.2 eV in the formation energy would expand the searching domain to some metastable candidate metal hydrides, which are only stable with a modest energy input. Similarly, the lower bound of the formation energy to -1.2 eV allows the inclusion of some more stable metal hydrides in the database that require mild energy to initiate hydrogen release. However, given these thermodynamic properties indeed account for the hydrogen storage and release, it seems unnecessary to strictly limit the formation energy of all candidate metal hydrides to less than or equal to zero. A slight relaxation of 0.2 eV on both bounds could still generate desired metal hydrides in real-world practice considering their elevated operating temperature/pressure beyond what are included in the database. For these reasons, the formation-energy weight factor is redefined as
\begin{equation}\label{eq:E_factor2}
E_{factor} = \begin{cases}
\sqrt{1 - \left(\frac{|E_{form} + 0.5|}{0.7}\right)^2}, & E_{form} \in [-1.2, 0.2] \text{ eV},\\
0, & \text{otherwise},
\end{cases} \end{equation}
with $E_{form}$ representing the formation energy. Notice that, since the range of $E_{form}$ has been expanded, the fraction used in the definition of $E_{factor}$ must be adjusted from 0.5 to 0.7 to maintain its value within the range of 0 to 1 (see Fig.~\ref{fig:E_factor}).

Such adjustments increase the pool of material candidates while still identifying those feasible for hydrogen storage applications under realistic operating conditions. Once the evaluation metric is finalized, we then apply the FCI algorithm to identify the features related to the hydrogen storage score.


\subsection{Fast Causal Inference (FCI) Algorithm}
\begin{figure}[ht]
\centering
\includegraphics[width=0.7\textwidth]{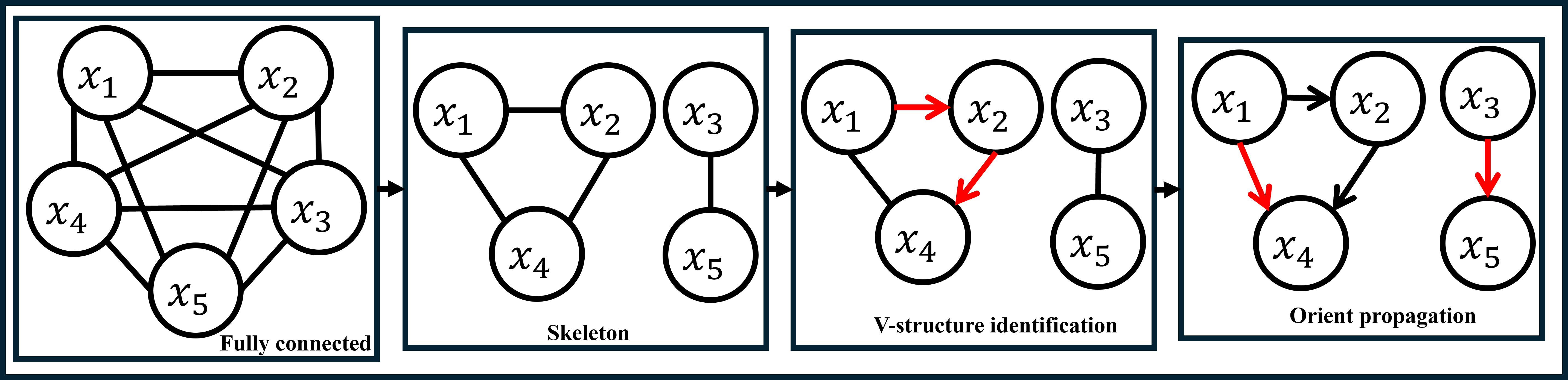}
\caption{Illustration of how the PC algorithm works.}
\label{fig:PC_Algo}
\end{figure}

To understand the FCI algorithm, it is helpful to first review the PC algorithm, as FCI extends its framework. The PC algorithm aims to learn the structure of a causal graph based on conditional independence relationships and consists of three key steps: the skeleton step, the v-structure identification step, and the orientation propagation step (see Fig.~\ref{fig:PC_Algo}).

It begins with a fully connected (i.e., complete) undirected graph and iteratively removes edges by conducting conditional independence tests, such as the Chi-square test for large datasets or Fisher’s Z test for smaller datasets (skeleton step). Once the graph skeleton is established, the algorithm identifies v-structures (i.e., colliders involving three nodes) using the results of the independence tests (v-structure identification step). Finally, the algorithm orients the remaining edges based on key assumptions of DGCMs, such as acyclicity and the causal Markov assumption (orientation propagation step).

Similar to the PC algorithm, the FCI algorithm follows the same three steps, starting with a fully connected undirected graph. Unlike the PC algorithm, the FCI algorithm assumes the presence of unobserved (or latent) confounders in the dataset. This allows the FCI algorithm to account for hidden relationships, making it generally more accurate at identifying causal structures than the PC algorithm.

In this study, we apply the FCI algorithm to identify features that are either directly related to or else causally influence hydrogen storage. Once the relevant features are determined, we then develop and train a VAE model to generate new potential material candidates.

\subsection{Variational Autoencoder (VAE)}

\begin{figure}[ht]
\centering
\includegraphics[width=0.7\textwidth]{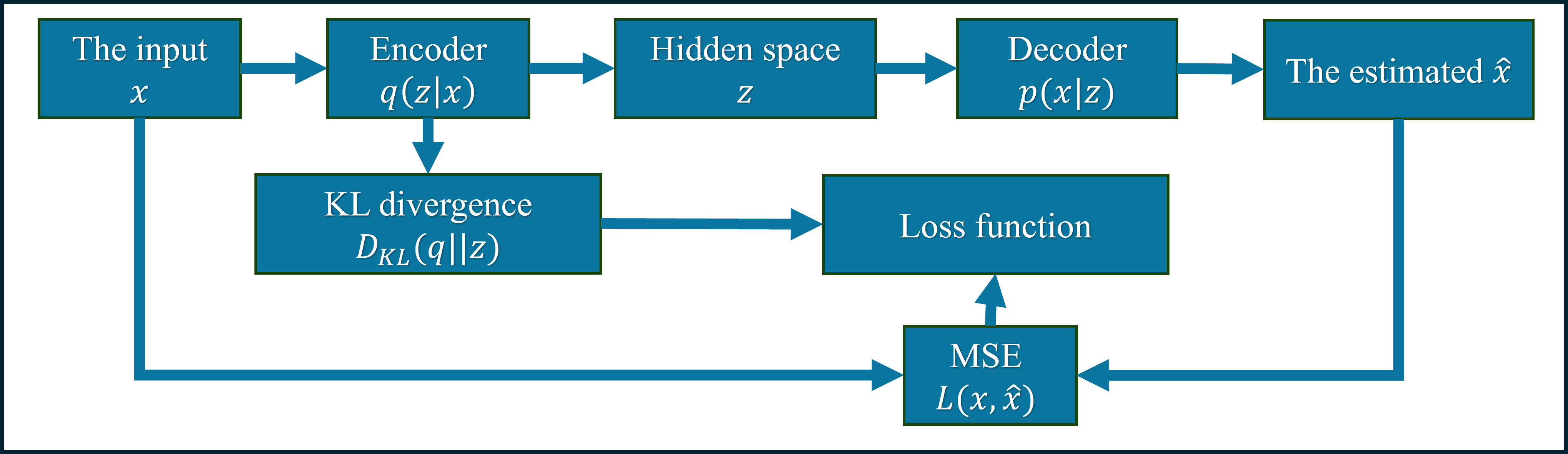}
\caption{Illustration of how the VAE algorithm works.}
\label{fig:VAE_demo}
\end{figure}

VAEs are generative models composed of two key components: an encoder and a decoder. The encoder captures the conditional distribution of the latent variables given the input, denoted as $q(z|x)$, while the decoder models the reverse process, $p(x|z)$, reconstructing the input $x$ from its latent representation $z$ (see Fig.~\ref{fig:VAE_demo}).

A central strength of VAEs lies in their ability to model and leverage unobserved latent variables within a probabilistic framework. The training objective involves minimizing a composite loss function that balances two competing goals: accurately reconstructing the input and regularizing the latent space. The first component of this loss function is the reconstruction loss, which is typically measured using Mean Squared Error (MSE) and evaluates how closely the decoder’s output matches the original input:
\begin{equation}
L_{MSE}(\theta, \phi) = \mathbb{E}[(x - f_{\theta}[g_{\phi}(x)])^2],
\end{equation}
where $g_{\phi}$ is the encoder that maps input $x$ to the latent space, parameterized by $\phi$, and $f_{\theta}$ is the decoder that maps from the latent space back to the reconstructed input, parameterized by $\theta$.

The second component of the loss function is the Kullback–Leibler (KL) divergence, which measures how much the learned latent distribution $q(z|x)$ deviates from a prior distribution $p(z)$, typically a standard normal distribution:
\begin{equation}
D_{\text{KL}}(q(z|x) || p(z)) = \sum_z q(z|x) \log\left(\frac{q(z|x)}{p(z)}\right).
\end{equation}
Combining both terms, the overall loss function of the VAE is given by:
\begin{equation}
L_{\text{VAE}} = L_{\text{MSE}} + D_{\text{KL}}(q(z|x) \parallel p(z)).
\end{equation}
This loss formulation enables the VAE not only to reconstruct inputs with high fidelity but also to enforce a well-structured latent space, which is essential for generating diverse, valid, and chemically meaningful candidates.

In this study, we apply the Crystal Diffusion Variational Autoencoder (CDVAE) \cite{xiecrystal}, a VAE-based model developed to generate crystal structures of chemical compounds. The CDVAE is trained using selected chemical formulas and their associated features from the Materials Project database \cite{Jain2013}, along with the computed hydrogen storage score defined in Equation \eqref{eq:HStorageScore}, which incorporates the modified formation energy factor from Equation \eqref{eq:E_factor2}. Once trained, the model is used to generate new candidate chemical formulas.

\subsection{M3GNet}
Although the CDVAE can theoretically generate crystal structures from chemical formulas, it may also produce structures that are energetically unstable or physically unrealistic. In extreme cases, such structures would release large amounts of energy if realized, making them impractical or unsafe. Therefore, the CDVAE alone is insufficient for generating physically stable, experimentally viable materials.

To address this, we refine their Crystallographic Information Files (CIFs) using M3GNet \cite{chen2022universal}. This GNN-based model is designed to relax crystal structures, providing a faster and more efficient alternative to traditional DFT simulations. M3GNet is well known for its accuracy and computational efficiency in predicting formation energy, achieving an mean absolute error (MAE) of 0.0754 eV and an $R^2$ value of 0.983. This refinement step helps verify that the generated materials and their crystal structures are suitable for practical hydrogen storage applications. It is worth noting that the performance of the proposed method, integrated with M3GNet, achieves an MAE of 0.0775 eV (see Table~\ref{tab:MSE} in Section~\ref{Experiments}).

\subsection{DFT calculation of formation energy}
The formation energy ($E_{form}$) is the energy change upon reacting to form a phase of interest from its constituent components. It is typically expressed as
\begin{equation}
E_{form} = E_{total} - \sum_{i} n_iE_i,
\end{equation}
where $E_{total}$ is the total energy of the phase of interest, $n_i$ is the number of moles of the $i$-th element, and $E_i$ is the energy of the $i$-th element in its reference state. In our calculations, the reference states of metal and non-metal elements are their stable solid phases while H reference state is the $H_2$ molecule.

DFT calculations were performed on each compound and $H_2$ molecules by employing Vienna \textit{AB-initio} Simulation Package (VASP) \cite{kresse1993ab, kresse1996efficient, kresse1996efficiency}. In the calculation, we use planewave basis set with a kinetic energy cutoff of 520 eV, the projector augmented wave (PAW) pseudo-potential to describe the electron–ion interactions, and the Perdew-Burke-Ernzerhof (PBE) scheme \cite{perdew1996generalized} to treat the exchange correction.

\section{Results and discussion}\label{Experiments}
\subsection{Hydrogen storage database and FCI algorithm result}
We first apply the FCI algorithm to the hydrogen storage database developed in a previous study \cite{nations2023metal}. This database contains 2,356 observations, each comprising 13 variables, including hydrogen storage score, band gap, density, and others. Among these variables, temperature and pressure exhibit limited variation, making them less suitable for further analysis. Using the FCI algorithm, we identify the closest causal neighborhood of the hydrogen storage score. These neighboring features are considered the most relevant and are used to determine the variables extracted from the Materials Project database for model training.

\begin{figure}[ht]
\centering
\includegraphics[width=0.7\textwidth]{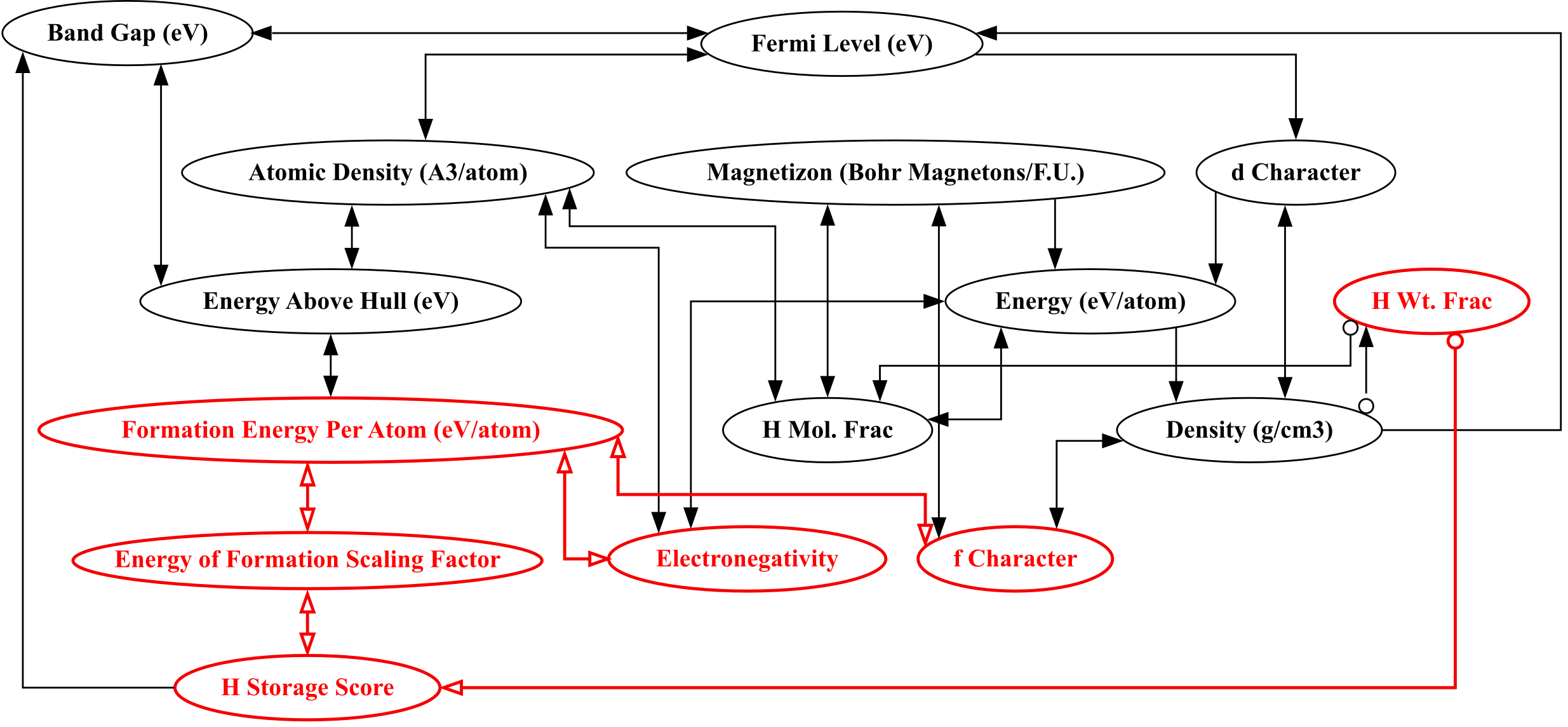}
\caption{Causal graph generated by the FCI algorithm using the Hydrogen Storage Database. Edges represent statistical dependencies identified through Chi-squared independence tests at a significance level of $\alpha = 0.05$. The focus is on the hydrogen storage score node, highlighting its relationship with key features such as hydrogen weight fraction ($W_{H_2}$), formation energy ($E_{form}$), Band Gap, and crystal structure (f Character). Bidirectional edges indicate correlation rather than causation.}
\label{fig:FCI_result}
\end{figure}

Fig.~\ref{fig:FCI_result} shows the results produced by the FCI algorithm. In this experiment, we apply the FCI method using the Chi-squared independence test with a significance level of $\alpha = 0.05$. The analysis is carried out with the Python package causal-learn~\cite{zheng2024causal}, which automatically discretizes continuous variables, such as density and band gap, before running the independence tests. Since the goal of this study is to identify the features that have the strongest association with the hydrogen storage score, we focus our analysis on the node representing this variable in the resulting causal graph.

The results suggest that the Band Gap is not a causal factor of the hydrogen storage score. Rather, changes in the hydrogen storage score appear to influence the Band Gap. Additionally, the node labeled H Wt Frac corresponds to the hydrogen weight fraction, denoted as $W_{H_2}$. According to the graph, once $W_{H_2}$ is known, the Density variable becomes redundant.

A similar conclusion can be drawn regarding the Energy of Formation Scaling Factor, denoted as $E_{factor}$. Along this path, all edges are bidirectional, indicating correlation rather than causation. It is also noteworthy that the f Character, a feature describing electron structure, is directly correlated with both Formation Energy ($E_{form}$) and Density, and is also closely related to $W_{H_2}$. From these observations, we conclude that the three most critical features influencing the hydrogen storage score are the hydrogen weight fraction ($W_{H_2}$), formation energy ($E_{form}$), and the crystal structure.

\begin{table}[ht]
  \centering
  \caption{Principal Component Regression (PCR) Models and Their Results. The independent variable is the hydrogen storage score, as defined in Equation \eqref{eq:HStorageScore}.}
    \begin{tabular}{cccccccc}
    \toprule
    \multicolumn{5}{c}{Dependent variables} &       & \multicolumn{2}{c}{Mean Squared Error (MSE)} \\
\cmidrule{1-5}\cmidrule{7-8} $E_{form}$ & Density & $W_{H_2}$ & Band Gap & other variables &       & Testing MSE & Training MSE \\
    \midrule
          &       & X     &       & X     &       & 3.4E-04 & 8.0E-05 \\
    X     & X     &       &       &       &       & 7.3E-04 & 3.4E-04 \\
    X     & X     & X     &       &       &       & \textbf{2.4E-04} & 6.0E-05 \\
    X     & X     &       & X     &       &       & 5.1E-04 & 2.2E-04 \\
    X     & X     &       & X     & X     &       & 5.6E-04 & 1.7E-04 \\
    X     & X     &       &       & X     &       & 7.3E-04 & 2.6E-04 \\
    X     & X     & X     & X     & X     &       & \textbf{2.7E-04} & 6.0E-05 \\
    \bottomrule
    \end{tabular}%
  \label{tab:PCRResult}
\end{table}

To further validate our findings, we perform a set of Principal Component Regression (PCR) models, as shown in Table~\ref{tab:PCRResult}. Notably, the model that includes both Band Gap and all other variables yields a higher testing Mean Squared Error (MSE) of $2.7 \times 10^{-4}$, compared to the model that includes only $E_{form}$, Density, and $W_{H_2}$, which achieves an MSE of $2.4 \times 10^{-4}$. These results, together with the output from the FCI algorithm, indicate that hydrogen weight fraction, formation energy, and crystal structure (represented by Density and f Character) are statistically sufficient to predict the hydrogen storage score. In other words, the hydrogen storage score is conditionally independent of all remaining variables (including latent confounders) given these features. This confirms that these variables are key criteria and sufficiently informative for the ML model to generate promising compounds for hydrogen storage.

Since the hydrogen storage score incorporates both $W_{H_2}$ and $E_{form}$, and density and f Character are directly related to crystal structure, we train the CDVAE using observations that include the chemical formula, the hydrogen storage score, and the corresponding CIF file, which represents the crystal structure of each compound. In this study, we filter the Materials Project database to prepare the training data for the CDVAE.

\subsection{Material project database and generative ML model}
To identify promising chemical formulas for hydrogen storage, we adopted a multi-step data-driven pipeline. First, we constructed a training dataset from the Materials Project database based on specific screening criteria. This dataset was then used to train the CDVAE model. After training, the CDVAE was used to generate a large set of candidate chemical formulas along with their corresponding crystal structures.

Next, we applied a set of domain-specific, rule-based filters to eliminate chemically implausible or irrelevant candidates. The filtered formulas were then passed to M3GNet to determine their relaxed crystal structures and recalculate the formation energy. Finally, we sorted the candidates by their modified hydrogen storage score and selected the top 100 most promising compounds for further analysis.

\subsubsection{Training dataset}
To train the CDVAE model, we selected a dataset from the Materials Project~\cite{Jain2013} database by applying the following selection criteria. These criteria are directly adopted from the original CDVAE paper~\cite{xiecrystal} to ensure stability, with the exception of the first criterion:
\begin{itemize}
\item The chemical formula must include hydrogen.
\item The crystal structure must have no more than 20 atomic sites. This threshold originates from ICSD~\cite{belsky2002new, xiecrystal} and helps ensure experimental stability of the identified candidates.
\item The energy above the hull must lie between 0 and 0.08~eV. This threshold originates from the CDVAE MP-20 constraints~\cite{ren2020inverse, xiecrystal}.
\item The formation energy must be less than or equal to 0~eV. This constraint ensures that the training dataset contains only thermodynamically stable compounds.
\end{itemize}

Based on these criteria, we obtain a total of 450 observations. Each observation includes three components: the chemical formula, the CIF file, and the original hydrogen storage score defined in Equation~\eqref{eq:HStorageScore}. To further optimize the method, the CDVAE includes a property optimization step. In this step, the inverse of the hydrogen storage score computed from the compounds generated by the decoder with a given random hidden space $z$ (see Fig.~\ref{fig:VAE_demo}) is used as an independent objective function that is minimized using gradient descent. The optimization is performed over 5,000 gradient steps with a step size of $1\times 10^{-3}$ \cite{xiecrystal}. This process helps guide the generative model toward materials with higher hydrogen storage performance. To ensure robustness, the dataset is split into training, validation, and testing sets at a 60:20:20 ratio, resulting in 270, 90, and 90 samples in the corresponding subsets.

\subsubsection{Filtering generated chemical formulas}

After training, the CDVAE model generates 1,000 candidate chemical formulas. To ensure these candidates are chemically realistic and potentially viable for hydrogen storage, we apply a series of domain-specific filtering rules based on chemical bonding theory and periodic trends. These rules remove physically impossible compounds, compounds that are stable but unsuitable for hydrogen storage and release, and economically costly compounds.

\begin{itemize}
\item Group 16 elements (chalcogens) can bond with at most two hydrogen atoms. This filter ensures that the generated candidates remain practically reasonable and excludes compounds such as $H_6S$ and $H_2S_2$, which are unstable and/or flammable, not suitable for hydrogen storage applications.
\item Group 15 elements (pnictogens) in the form $N_nH_{n+2}$ can bond with up to $n+2$ hydrogen atoms. This constraint helps ensure that the generated candidates are physically stable and suitable for hydrogen storage applications.
\item Group 14 elements can form compounds like $N_nH_{2n+2}$, $N_nH_{2n}$, or $N_nH_2$, depending on bonding configurations, allowing up to $2n+2$ hydrogen atoms.
\item Group 13 elements typically form compounds of the form $N_nH_m$, where $m < 2n$.
\item Metals and metal alloys can accommodate varying numbers of hydrogen atoms, with $N_nH_{2n}$ being the most common stoichiometry.
\item Exclude formulas composed solely of N, O, S, P, Se, and H, as they are unlikely to exhibit the desired metallic bonding characteristics.
\item Exclude formulas composed only of non-metals and hydrogen.
\item Include only formulas that contain at least one metal element.
\item Exclude any formula containing Group 17 (halogen) elements.
\end{itemize}
To improve the efficiency of subsequent DFT validation, we further restrict consideration to compounds composed of three or four distinct elements plus hydrogen.

\subsubsection{Feature recalculation}
After applying the filtering criteria to the CDVAE-generated chemical formulas, M3GNet is employed to relax the resulting structures. Following relaxation, the formation energy ($E_{form}$), hydrogen weight fraction ($W_{H_2}$), and modified hydrogen storage score are recalculated using the updated structural information.

\begin{table}[ht]
  \centering
  \caption{First 10 potential candidates generated by the proposed method. “Same Formula” indicates an exact match with a compound in the MP database, while “Same Ratio” indicates matching elemental ratios but not an identical formula.}
  \label{tab:first10obs}
  \resizebox{\columnwidth}{!}{
    \begin{tabular}{lccccccc}
    \toprule
    \multicolumn{1}{p{6.5em}}{Formula} & 
    \multicolumn{1}{p{7.0em}}{\centering $E_{form}$ Predicted by proposed method (eV/atom)} & 
    \multicolumn{1}{p{7.0em}}{\centering $E_{form}$ calculated by DFT (eV/atom)} & 
    \multicolumn{1}{p{3.5em}}{\centering H Storage Score} & 
    \multicolumn{1}{p{3.5em}}{\centering Squared error} &
    \multicolumn{1}{p{4.5em}}{\centering MP Unique ID} & 
    \multicolumn{1}{p{3.5em}}{\centering Same Formula} & 
    \multicolumn{1}{p{3.0em}}{\centering Same Ratio} \\
    \midrule
    Li$_3$B$_3$H$_6$ & 0.057  & -0.189  & 0.062  & 0.061  & mp-568523 & FALSE & TRUE \\
    Li$_1$Al$_3$H$_6$ & 0.019  & 0.049  & 0.043  & 0.001  &       & FALSE & FALSE \\
    \multirow{2}[0]{*}{Ti$_1$H$_2$} & \multirow{2}[0]{*}{-0.426 } & -0.466  & \multirow{2}[0]{*}{0.040 } & \multirow{2}[0]{*}{0.002 } & \multirow{2}[0]{*}{mp-1077482} & \multirow{2}[0]{*}{TRUE} & \multirow{2}[0]{*}{TRUE} \\
          &       & -0.448 \textsuperscript{a} &       &       &       &       &  \\
    \multirow{2}[0]{*}{Ti$_2$H$_4$} & \multirow{2}[0]{*}{-0.401 } & -0.465  & \multirow{2}[0]{*}{0.040 } & \multirow{2}[0]{*}{0.004 } & \multirow{2}[0]{*}{mp-1077482} & \multirow{2}[0]{*}{FALSE} & \multirow{2}[0]{*}{TRUE} \\
          &       & -0.401 \textsuperscript{a} &       &       &       &       &  \\
    K$_2$Al$_3$H$_6$ & -0.159  & 0.080  & 0.032  & 0.057  &       & FALSE & FALSE \\
    Ti$_6$H$_8$ & -0.387  & -0.315  & 0.027  & 0.005  &       & FALSE & FALSE \\
    Ti$_3$H$_4$ & -0.322  & -0.171  & 0.026  & 0.023  &       & FALSE & FALSE \\
    Ti$_5$H$_6$ & -0.361  & -0.350  & 0.024  & 0.000  &       & FALSE & FALSE \\
    Ca$_2$Al$_1$Si$_2$H$_3$ & -0.404  & -0.243  & 0.018  & 0.026  &       & FALSE & FALSE \\
    Ti$_4$Ni$_1$H$_4$ & -0.390  & -0.372  & 0.016  & 0.000  &       & FALSE & FALSE \\
    \bottomrule
    \end{tabular}}

\vspace{1ex}
\begin{minipage}{0.95\linewidth}
\small
\textsuperscript{a} Taken from Materials Project.
\end{minipage}
\end{table}%

\subsubsection{Outcomes and validation}
In this study, we apply the proposed method to generate a list of 1,000 candidate materials, which are then ranked in descending order according to their H Storage Score. From this list, the top 100 candidates are selected and added to the database for further analysis. Table \ref{tab:first10obs} displays the first 10 generated candidates, along with their corresponding information from the database and the DFT-calculated formation energy for comparison. The Squared Error is calculated as:
\begin{equation}
\text{Squared Error} = (\hat{E}_{form} - E_{form})^2,
\end{equation}
where $\hat{E}_{form}$ is the formation energy predicted by the proposed method, and $E_{form}$ is the formation energy calculated by DFT. In this study, the Mean Squared Error (MSE) of the model, calculated as the average of all squared prediction errors shown in the table, is 0.018, and the MAE is approximately 0.157. It is important to note that all these chemical compounds and their corresponding crystal structures are generated by the proposed model and are therefore unobserved in the MP database, making them unsuitable for directly comparing the performance of the M3GNet model.

Note that several candidates with the same chemical formula may have different crystal structure arrangements; the table shows only the most likely structure for each candidate. Additionally, each candidate is compared with compounds available in the Materials Project database and classified into three categories: (1) an exact match in chemical formula, (2) the same elemental ratio but a different total number of atoms, and (3) the same combination of elements, regardless of their number or ratio.

\begin{figure}[ht]
\centering
\includegraphics[width=0.7\textwidth]{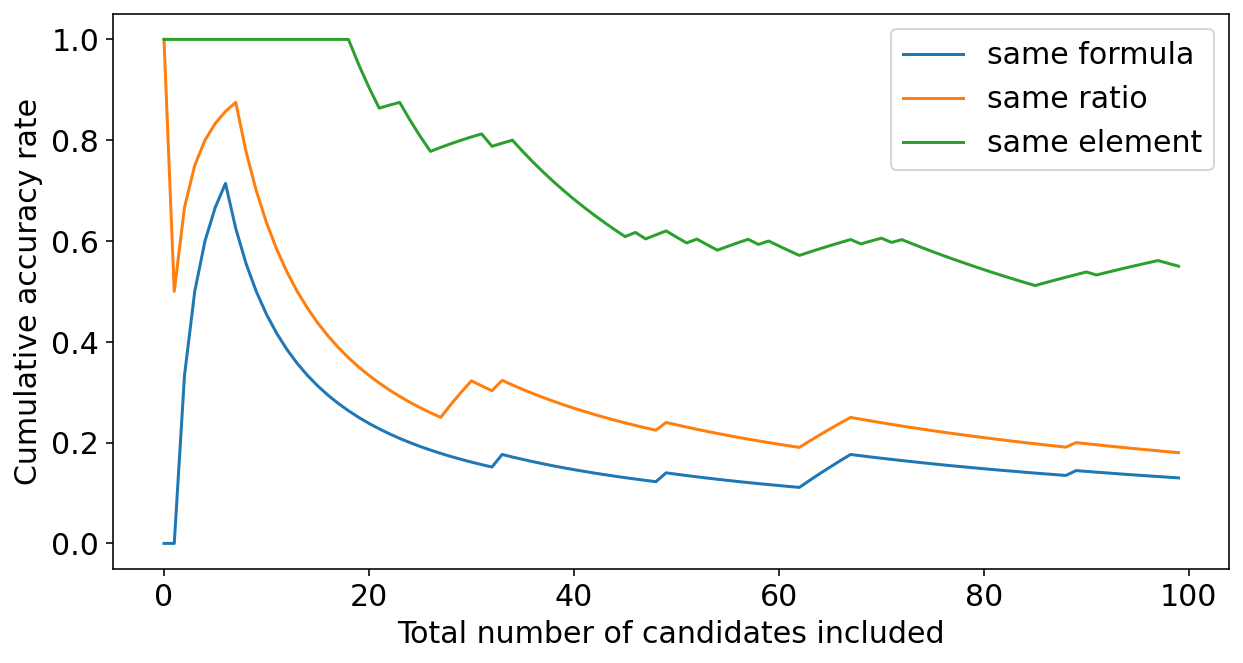}
\caption{Cumulative accuracy rate for the three evaluation criteria: “same formula” indicates an exact match with a compound in the MP database, “same ratio” indicates matching elemental ratios but not the exact formula, and “same element” indicates that only the elemental composition overlaps.}
\label{fig:AccuracyRate}
\end{figure}

Fig.~\ref{fig:AccuracyRate} shows the cumulative accuracy rates for the three matching categories. Candidates are ranked from the highest to lowest hydrogen storage score, and the accuracy rate is calculated based on the top $n$ candidates, where $n = 1,2,...,100$. As candidates with lower scores are included, the accuracy rate decreases, suggesting that the hydrogen storage score is a reliable metric for identifying promising hydrogen storage materials. Among the top 20 candidates, the accuracy rates for exact formula matches, same elemental ratios, and same elemental combinations are 0.25, 0.35, and 0.95, respectively.

Overall, the DFT calculated formation energies are very close to our predicted values, as well as to those reported in the Materials Project. The exceptional cases are $K_2Al_3H_6$ and $Li_3B_3H_6$. In the literature, we didn’t find the crystal structures of these two compounds. For $K-Al-H$, we found three compounds in the Materials Project database: $KAlH_3$, $KAlH_4$, and $K_3AlH_6$.
The DFT calculation of the formation energy indicates that $K_2Al_3H_6$ may not be stable, whereas our ML prediction indicates it is stable. For the $Li_3B_3H_6$ case, in the Materials Project database, we found $LiBH_2$, which could serve as the building block of $Li_3B_3H_6$, as there is a factor of 3 in their formulas. Our DFT-calculated formation energy indicates that $Li_3B_3H_6$ could be stable, whereas our ML prediction suggests otherwise. To further validate the stability of these two compounds, experimental synthesis and measurements are highly recommended.

\begin{table}[ht]
  \centering
  \caption{Comparison of predicted formation energies using the proposed method with observed formation energies from the Materials Project database.}
	\label{tab:MSE}
	\resizebox{\columnwidth}{!}{
    \begin{tabular}{lcccccc}
	\toprule
    \multicolumn{1}{p{7.19em}}{Formula} &
    \multicolumn{1}{p{4.75em}}{\centering Predicted $E_{form}$ (eV/atom)} &
    \multicolumn{1}{p{6.75em}}{\centering Materials Project $E_{form}$ (eV/atom)} &
    \multicolumn{1}{p{3.375em}}{\centering Squared Error} &
    \multicolumn{1}{p{5.94em}}{\centering MP Unique ID} &
    \multicolumn{1}{p{6.5em}}{\centering Crystal System} &
    \multicolumn{1}{p{3.75em}}{\centering Same Formula} \\
	\midrule
    B$_4$H$_5$ & 0.6603  & -0.0813  & 0.5499  & mp-1183667 & Monoclinic & TRUE  \\
    Ba$_1$Al$_1$Si$_1$H$_1$ & -0.4334  & -0.4598  & 0.0007  & mp-571093 & Trigonal & TRUE  \\
    Ba$_1$Ga$_1$Sn$_1$H$_1$ & -0.5353  & -0.5414  & 0.0000  & mp-1018094 & Trigonal & TRUE  \\
    Li$_3$B$_3$ H$_6$ & 0.0574  &  -0.1888\textsuperscript{a}& 0.0606 & NA &       & FALSE \\
    Ti$_1$H$_2$ & -0.4264  & -0.4479  & 0.0005  & mp-1077482 & Tetragonal & TRUE  \\
    Ti$_1$Ni$_1$H$_1$ & -0.2306  & -0.3351  & 0.0109  & mp-1071458 & Orthorhombic & TRUE  \\
    Ti$_2$H$_1$ & -0.2727  & -0.2278  & 0.0020  & mp-1077045 & Cubic & TRUE  \\
    Ti$_2$H$_4$ & -0.4006  & -0.4652\textsuperscript{a} & 0.0042 & NA &       & FALSE \\
    Ti$_4$H$_3$ & -0.3039  & -0.3099  & 0.0000  & mp-1078123 & Tetragonal & TRUE  \\
    Ti$_4$H$_3$Pd$_2$ & -0.4096  & -0.3288  & 0.0065  & mp-1080554 & Tetragonal & TRUE  \\
	\bottomrule
    \end{tabular}}%

\vspace{1ex}
\begin{minipage}{0.95\linewidth}
\small
\textsuperscript{a} Materials Project didn't have this chemical compound. Calculated by DFT.
\end{minipage}
\end{table}%

Since the proposed method generates candidates with formulas that match those in the Materials Project database, we can compare the predicted formation energy with the observed formation energy from the database. Table \ref{tab:MSE} lists all candidates that share the same formula with entries in the Materials Project database. Note that some formulas only share the same elemental ratio, not the exact formula; therefore, their formation energies are not available in the database and are replaced by DFT-calculated values instead.

The MSE for the results shown in Table~\ref{tab:MSE} is approximately 0.064. For comparison, the MAE is approximately 0.0775, which is close to the value reported for the original M3GNet model trained on the full Materials Project database of 188,000 structures (0.0754). This outcome demonstrates strong performance, especially considering that the hydrogen storage dataset used in this study is significantly smaller (270 training samples, 90 validation samples, and 90 testing samples) and therefore offers substantially less training diversity than the full MP dataset.

\section{Conclusion and future study}\label{Conclusion}
\subsection{Conclusion}
In this study, we propose a comprehensive framework that performs three main tasks: evaluation, feature discovery, and candidate generation to identify potential hydrogen storage materials that are unobserved in existing databases.


In summary, the following contributions were made:
\begin{itemize}
\item Develops a generative modeling framework that overcomes the limitations of traditional discriminative ML models by enabling the direct generation of new chemical formulas and candidate materials for hydrogen storage.
\item Proposes a lightweight CDVAE-based generative model trained on only 270 samples, enabled by selecting the most important features using the FCI algorithm.
\item Proposes a modified hydrogen storage score that expands the feasible design space and provides a more practical assessment of material performance.
\item Four entirely new alloy hydrides, not found in any existing materials databases, are discovered and validated through DFT simulations, demonstrating strong hydrogen storage potential and suitability for future experimental study.

\end{itemize}



\subsection{Future study}
The current model still presents several limitations. The first limitation arises from the causal structure identified by the FCI algorithm (see Fig.~\ref{fig:FCI_result}), which incorrectly suggests that the hydrogen storage score is a causal parent of the band gap. This result contradicts established physical principles and occurs because causal discovery approaches such as FCI rely entirely on statistical patterns and graphical assumptions rather than domain knowledge. Although this issue does not affect the workflow in this study, since the causal graph is used only for feature selection (see Table~\ref{tab:PCRResult}) before the ML modeling step, it indicates the need for improved interpretability. Future work may address this limitation by incorporating functional causal models that allow constraints informed by physical reasoning~\cite{wang2024survey}.

Another limitation concerns the flexibility of the hydrogen storage score defined in Equation~\eqref{eq:HStorageScore}. In certain cases, such as $\text{Li}_1\text{B}_4\text{H}_7$~\cite{yang2024application}, materials with relatively high hydrogen weight fraction $W_{H_2}$ may still receive a low storage score when the formation energy $E_{\text{form}}$ is very low. This behavior suggests that the current scoring formulation may underestimate the performance of some promising materials and should be further refined.
One potential extension is to integrate a GAN-based framework in which the CDVAE acts as the generator and an improved adaptive scoring metric functions as the discriminator to evaluate physical plausibility, stability, and practical feasibility.

Finally, although the characterization of the structural properties of metal hydrides and their hydrogen storage capacity has been well advanced by our work and others, the time-scale behaviors associated with hydrogen release and the accompanying evolution of crystal structures remain largely unexplored \cite{wei2022toward, yu2024hydrogen}. Studies that integrate molecular dynamics simulations with time-series ML models, such as LSTM-based models or attention-based generative modeling approaches, may provide effective tools for capturing temporal behavior and structural evolution in metal hydride systems. Such ML based predictions of dynamic behavior or kinetics of hydrogen release represent new research directions that are worthy of further investigation.




\section{Author Contributions: CRediT}
\textbf{Xiyuan Liu:} Principal Investigator, Data Analysis, Supervision, Writing (Original Draft).
\textbf{Christian Hacker:} Investigator, Data Analysis, Programming, Writing (Review and Editing).
\textbf{Shengnian Wang:} Supervision, Resources, Writing (Review and Editing).
\textbf{Yuhua Duan:} Conceptualization, Resources, Supervision, Project Administration, Writing (Review and Editing).

\section{Acknowledgment statement}
This work is supported by the US National Science Foundation under grant number OIA-1946231 and the Louisiana Board of Regents for the Louisiana Materials Design Alliance (LAMDA).

Y.D. was supported in part by the National Energy Technology Laboratory (NETL) Research and  Innovation Center’s Advanced Sensors, Control, and Novel Concept Program (MYRP\#1025037).

\section{Data availability}
The data presented in this study are available on request from the corresponding author due to they are part of several ongoing research projects.

\section{Declaration of competing interest}
The authors declare that they have no known competing financial interests or personal relationships that could have appeared to influence the work reported in this paper.

\bibliographystyle{elsarticle-num-names}
\bibliography{references}  






\end{document}